\documentclass[10pt,twocolumn,letterpaper]{article}

\usepackage{iccv}
\usepackage{times}
\usepackage{epsfig}
\usepackage{graphicx}
\usepackage{amsmath}
\usepackage{amssymb}
\usepackage{booktabs}
\usepackage{xcolor}
\usepackage{pifont}
\usepackage{caption}

\usepackage[pagebackref=true,breaklinks=true,letterpaper=true,colorlinks,bookmarks=false]{hyperref}

\iccvfinalcopy 

\def\iccvPaperID{***} 

\newcommand{\New}[1]{#1}
\newcommand{\xmark}{\ding{55}}
\newcommand{\cmark}{\ding{51}}
\begin{document}

\title{Learning Space-Time Semantic Correspondences}

\author{Du Tran\\
Samsung Research America \and
Jitendra Malik\\
UC Berkeley
}

\ificcvfinal\thispagestyle{empty}\fi

\twocolumn[{%
\renewcommand\twocolumn[1][]{#1}%
\maketitle
\begin{center}
    \centering
    \captionsetup{type=figure}
    \includegraphics[width=\textwidth]{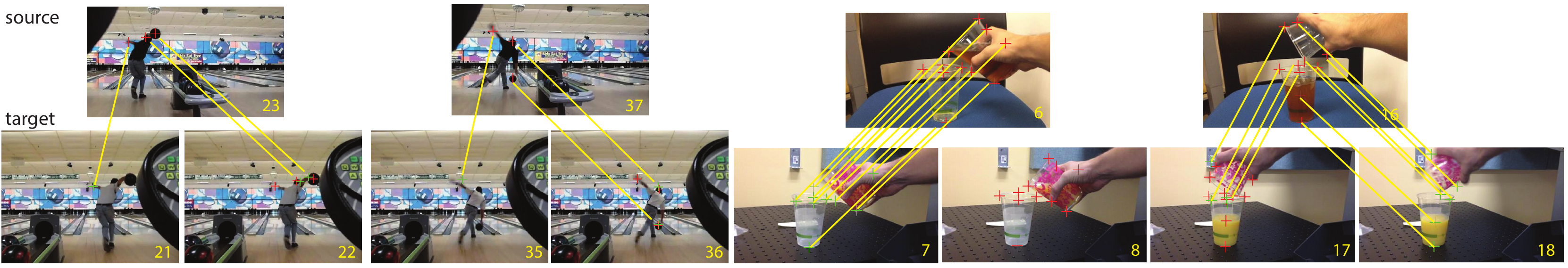}
    \caption{{\bf Space-time semantic correspondence prediction}. Two examples of the space-time semantic correspondence prediction problem are visualized with annotations and predictions. The upper row shows the source video frames while the lower one presents the target sequences (frame numbers are indicated at the lower right corners). All ground-truth keypoints are visualized in red and predicted keypoints are in green markers. The space-time semantic correspondence prediction problem requires the pair of correspondences to be aligned in both space and time. In these examples, the time alignment happened at key moments: ``ball swung fully back'' and ``ball release'' for bowling videos, or ``pouring start'' and ``pouring end'' for pouring videos. The matching predictions are visualized with yellow lines. The keypoints in the 21st, 35th frames of the bowling and the 7th, 18th frames of the pouring target videos are misaligned by 1 frame. Best viewed in color.}
    \label{fig:intro}
\end{center}%
}]

\begin{abstract}
We propose a new task of space-time semantic correspondence prediction in videos. Given a source video, a target video, and a set of space-time key-points in the source video, the task requires predicting a set of keypoints in the target video that are the semantic correspondences of the provided source keypoints. We believe that this task is important for fine-grain video understanding,   potentially enabling applications such as activity coaching, sports analysis, robot imitation learning, and more. Our contributions in this paper are: (i) proposing a new task and providing annotations for space-time semantic correspondences \New{on two existing benchmarks: Penn Action and Pouring}; and (ii) presenting a comprehensive set of baselines and experiments to gain insights about the new problem. Our main finding is that the space-time semantic correspondence prediction problem is best approached jointly in space and time rather than in their decomposed sub-problems: time alignment and spatial correspondences. 

\end{abstract}


\section{Introduction}
\label{sec:intro}

{\bf What are space-time semantic correspondences?} Given two videos $\mathcal{V}_{1}$ and $\mathcal{V}_{2}$ which are assumed to have similar semantic content, \eg, two videos of people performing the same actions. The two space-time keypoints $p:(x_p,y_p,t_p)$ in $\mathcal{V}_{1}$ and $q:(x_q,y_q,t_q)$ in $\mathcal{V}_{2}$ are defined as the space-time semantic correspondence of each other when they are semantically aligned in both space and time. More specifically, $p$ and $q$ are semantically aligned in time when $t_p$ and $t_q$ are the correct alignment of each other defined by the key moments~\cite{Dwibedi_2019_CVPR} in $\mathcal{V}_{1}$ and $\mathcal{V}_{2}$. And $p$ and $q$ are semantically aligned in space when $(x_p,y_p)$ and $(x_q,y_q)$ are the true visual semantic correspondence of each other at the $t_p$-th frame of $\mathcal{V}_{1}$ and the $t_q$-th frame of $\mathcal{V}_{2}$.

{\bf Space-time semantic correspondence prediction}. Given a pair of videos: a source video $\mathcal{V}_{S}$ and a target video $\mathcal{V}_{T}$, and a set of space-time keypoints $P_S$ in $\mathcal{V}_{S}$, the problem of \emph{space-time semantic correspondence prediction} is to predict the set of keypoints $P_T$ in $\mathcal{V}_{T}$ those are the space-time semantic correspondences of $P_S$. Figure~\ref{fig:intro} provides two examples of space-time semantic correspondence prediction in two pairs of videos: one includes ``bowling'' videos, and the other includes ``pouring'' videos. Ground truth space-time semantic correspondences are visualized with red markers in these videos. The ground truth keypoints are temporally aligned by key moments: ``ball swung fully back'' and ``ball release'' for bowling, or ``liquid starts pouring'' and ``liquid stops pouring'' for pouring videos. The ground truth keypoints are also spatially aligned at semantic keypoints: head, wrists, bowling ball (in bowling videos) and fingertips, cup corners, and hand (in pouring videos).

{\bf Why's this problem important?} This problem, if solved, will enable various practical applications including activity coaching, sports analysis, and robot imitation learning. In activity coaching, a space-time semantic correspondence prediction model may assist to point out the differences between a professional golf player versus a novice. The model can also be useful in assessing how well a person is performing the bowling swing compared with herself or himself one month ago. In sports analysis, a similar model can be used to analyze and compare different players and provide feedback. In robot imitation learning, a robot may watch the human teacher in an exo-view while it imitates the task in an ego-view. The space-time semantic correspondence prediction can also be adopted to solve the correspondence matching across ego-exo views. In addition to that, we believe the problem of matching space-time keypoints semantically across videos and views is fundamental as models are required to understand the key moments, objects, and their interactions to complete the task.

Our contributions in this paper are: 
\begin{itemize}
    \item We propose a novel task of space-time semantic correspondence prediction which is an essential task for video understanding with various practical applications.
    \item We provide {\bf two} new datasets for this task by adding space-time semantic keypoint annotations to two existing datasets: Penn Action~\cite{penn_action} \New{and Pouring~\cite{PouringDataset}}.
    \item We present a set of comprehensive baseline approaches and perform an in-depth analysis to gain insights about the new problem. All annotations, source code, and models will be released upon publication.
\end{itemize}

\section{Related Work}
\label{sec:related_work}

{\bf Visual semantic correspondences in images}. Visual semantic correspondence prediction in images is a fundamental problem and well-studied~\cite{TaniaiSS16,ham2016,min2019spair}. Early methods approached this problem by local desctiptor matching~\cite{liu08,kim2016,Bristow15,cho15,ham2016,TaniaiSS16} normally with hand-crafted features, \eg, ~SIFT~\cite{sift} or HOG~\cite{hog}. After deep learning, CNN features are also used for semantic correspondence matching~\cite{long14,choy16,han17,kim17,kim17b}. More recently, visual semantic correspondence prediction is approached by various architecture-based methods including Hyperpixel~\cite{min2019hyperpixel}, \New{Neighbourhood Consensus Networks~\cite{NCNet}, Multi-scale Matching Networks~\cite{MMNet}}, Optimal Transport problem~\cite{liu2020semantic}, Dynamic Hyperpixel Flow~\cite{min2020dhpf}, Convolutional Hough Matching~\cite{min2021chm}, Cost Aggregation Transformers (CATs)~\cite{cho2021cats,cho2022cats_plus_plus}, Volumetric Aggregation with Transformers (VATs)~\cite{hong2022cost}. Inspired by the fundamental and practical nature of this problem in the image domain, we extend this problem into space-time and study the extended problem in videos.

{\bf Time alignment in videos}. Although time alignment is well studied in time-series analysis~\cite{dtw,soft_dtw}, there are not many works on video time alignment. Cao~\etal~\cite{cao20} used video time alignment for few-shot video classification. Yi~\etal~\cite{yi19} utilized video transcript alignment for weakly-supervised learning. More recently, dense temporal alignment in videos is used for self-supervised learning~\cite{Dwibedi_2019_CVPR,Haresh21}. The later work~\cite{Dwibedi_2019_CVPR,Haresh21} are closely related to ours, however, their problem setup is dense temporal alignment and ignores the spatial details. In contrast, our problem is set up to perform space-time alignment and at only sparse space-time keypoints, \eg, at semantic keypoints in key-moment frames.

\New{
{\bf Space-time correspondences in videos}. Space-time correspondences have been previously studied in videos. Wang~\etal~\cite{XiaolongWangCVPR19} proposed cycle consistency in time for visual image representation learning. Jabri~\etal~\cite{jabri2020walk} further employed random walks and cycle consistency for self-supervised learning. More recently, Son~\cite{SonCVPR22} proposed a contrastive learning approach using self-cycle consistency for self-supervised representation learning. We note that these works are self-supervised learning methods that utilize space-time correspondences within the same video to learn visual representations. In contrast, our work is a supervised-learning approach that predicts space-time correspondences across two different videos and for predicting semantic correspondences as opposed to learning visual representations.
}

{\bf Cross-video semantic prediction}. The Action Similarity Labeling Challenge (ASLAN)~\cite{ASLAN} is also related to our work in terms of cross-video semantic labeling where models have to predict if two input videos contain the same semantic action or not, \eg, both videos of playing soccer. Different from ASLAN, our problem requires models to predict semantic correspondences across videos at the keypoint level, not just at the action level.

\section{Benchmark Construction}
\label{sec:benchmark}

In this work, we adopt two existing benchmarks: Penn Action~\cite{penn_action} and Pouring~\cite{PouringDataset} for our new task of space-time semantic correspondence prediction. The following subsections describe the process for annotating these benchmarks. 

\subsection{Penn Action}

{\bf Data selection}. Penn Action was proposed for action recognition which contains 2,326 videos of 15 human actions, and all video frames are provided with 2D human keypoints. Penn Action is suited for our study because it was previously used in time alignment problem~\cite{Dwibedi_2019_CVPR} and provided with 2D human keypoint annotations which we can use as space-time keypoints. Since our problem requires aligning the keypoints both in space and time, we also adopt the definition of key moments for Penn Action used in~\cite{Dwibedi_2019_CVPR} for ``semantic'' time alignment. As we are interested in aligning space-time keypoints that capture both the subjects and the objects involved in the actions, \eg, bowling requires interacting with a bowling ball or playing golf requires using a golf stick and a ball, we eliminate actions involving no object such as ``jumping jacks'', ``Pushups'', ``Situps''. We also eliminate actions that have only one key moment, \eg, ``bench press'' and ``Pullups'' as it requires less time alignment. Table~\ref{tab:st_keypoints} presents the selected actions with their associated key moments and the objects involved during these actions. 

\begin{table*}[t]
\centering
{\small
\begin{tabular}{|l|l|l|}
        \hline
        {\bf Action} & {\bf Key moments} & {\bf Involving objects} \\
        \hline
        Baseball pitch & Knee fully up, Arm fully stretched out, Ball release & ball\\
        Baseball swing & Bat swung back fully, Bat hits ball & bat, ball \\
        Bowling & Ball swung fully back, Ball release & ball \\
        Golf swing & Stick swung fully back, Stick hits ball & golf stick, ball \\
        Squats & Hips at knees (going down), Hips at floor, Hips at knee (going up) & bar (center), left- and right discs \\
        Tennis forehand & Racket swung fully back, Racket touches ball & racket, ball \\
        Tennis serve & Ball released from hand, Racket swung fully back, Ball touches racket & racket, ball\\
        \hline
\end{tabular}%
\vspace{-8pt}
\caption{{\bf Selected actions with key moments and objects}. We select a subsect of actions from Penn Action~\cite{penn_action} for annotating our benchmark. Beside adopting the key moment definitions from~\cite{Dwibedi_2019_CVPR}, we define the set of objects that are involved in each action for annotating.}
\label{tab:st_keypoints}
}

\end{table*}

{\bf Annotating space-time keypoints}. We define our space-time semantic keypoints as 2D semantic keypoints happening at the key moments (the second column of Table~\ref{tab:st_keypoints}). A 2D semantic keypoint is a spatial location in the image which has its own semantic meaning that can be matched with a similar 2D semantic keypoint in another image. Examples of semantic keypoints can be human or object keypoints such as the left knee, the right wrist, the head of a person, a golf stick, or a bowling ball, etc. By this definition, we can leverage the human keypoints from Penn Action at the key-moment frames as our space-time semantic keypoints. Since we need also semantic keypoints on the objects, we annotate the keypoints for involving objects (the last column of Table~\ref{tab:st_keypoints}) at the key-moment frames whenever they are visible. \New{To ensure consistency in the object keypoints, we explicitly define the object keypoints as follows. For circular objects such as tennis ball, gofl balls, baseball ball, bolwing balls, and gym discs, the keypoints are the center of these objects. For bar-shaped objects such as baseball bats and gym bars, the keypoint is at the center of the bar. For golf sticks, the keypoint is at the club-head, and for tennis rackets, the keypoint is at the center of the racket head.}

{\bf Constructing pairs of correspondences}. Because we have all human and object semantic keypoints annotated at the key moments in each video, in theory, any pair of videos with the same action label can be used to form a pair for our task. This can be done by selecting one video as the source and the other as the target, and using space-time semantic keypoints in these two videos as space-time semantic correspondences of each other. In practice, not all key-points at key-moment frames are visible, thus we can only form a pair when two videos (with the same action label) share a minimum number of visible keypoints (\eg, $3$). We also present two different benchmark setups for our problem (Table~\ref{tab:benchmark_setup}). The ``13+3'' setup uses all semantic keypoints available which could be up to 13 human keypoints and up to 3 object keypoints per frame. The ``3+3'' setup is designed to balance between human keypoints and object keypoints. 

\begin{table}
\centering
{\small
\begin{tabular}{|c|l|c|c|c|}
        \hline
        setup & selected human keypoints & objects & \# train & \# val \\
        & & & pairs & pairs \\ 
        \hline
        13+3 & All 13 human keypoints & all & 39.9k & 21,6k\\
        3+3 & head, left and right wrists & all & 39.9k & 21,5k\\
        \hline
\end{tabular}%
\vspace{-8pt}
\caption{{\bf Benchmark setup \& statistics}. We experiment with two benchmark setup for space-time semantic correspondences. The 13+3 uses all human keypoints from Penn Action (up to 13) and all object keypoints (up to 3 per frame). The 3+3 uses only 3 key-points on human (head, left and right wrists) and all object keypoints.}
\label{tab:benchmark_setup}
}
\end{table}

{\bf Benchmark size and split}. We annotated object keypoints for all 1,482 videos of the actions listed in Table~\ref{tab:st_keypoints}. We use the same training and validation splits defined in the original Penn Action dataset, meaning using only video in the training split to form the pairs for our training split (similarly for the validation split). Even though the number of annotated videos is moderate, the number of pairs is much larger. The numbers of training and validation pairs are shown in Table~\ref{tab:benchmark_setup} (the 3+3 setup has a lightly smaller number of pairs as a result of removing pairs with fewer than 3 visible keypoints). An example of bowling in our dataset (3+3 setup) is visualized with both ground truth and predicted keypoints in Figure~\ref{fig:intro}.

\begin{figure}[!tbp]
  \centering
  \includegraphics[width=0.4\textwidth]{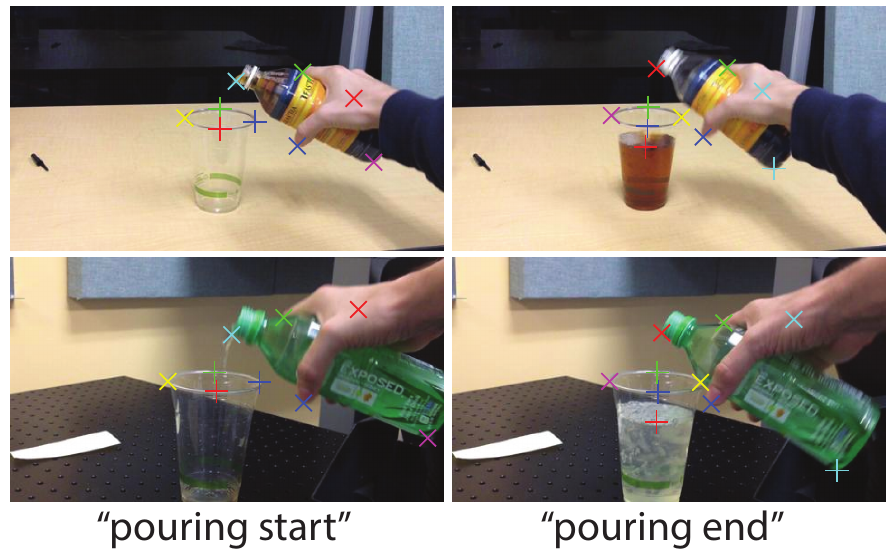}
  \vspace{-8pt}
  \caption{{\bf An example of pouring}. The upper row shows the source video, and the lower row shows the target video. The two key moment frames are visualized with key-points. Each pair of corresponding key-points is visualized by markers with the same type and color.}
  \label{fig:pouring}
\end{figure}

\New{
\subsection{Pouring}
The Pouring dataset is proposed and used in robotic research~\cite{PouringDataset,SermanetXL17} which includes 17 (11 training, 6 testing) videos of a human hand pouring liquid from a container into a cup. We define key moments in pouring as the time when the liquid starts and stops pouring. Since this dataset is quite small, we don't need to pre-define a fixed set of spatial semantic keypoints for annotating. Indeed, we can annotate each video pair independently to maximize the keypoint diversity. Our annotation process is described as follows. First, we annotate the pre-defined key moments for each video. Next, for each pair of videos, we annotate each pair of frames at the key moments independently. The keypoints are normally selected at the center of the hand, the fingertips, the corners of the liquid container, and the corners of the cup. This annotation process provides us with 55 training and 15 testing pairs of pouring videos. Fig~\ref{fig:pouring} shows one example from the Pouring dataset with annotations.}

\New{
{\bf Verification and correction}. For both Penn Action and Pouring, after the annotation finishes, all key-moment frames, visualized with annotated keypoints, are shown to annotators for quality assessment and potential corrections.}

\section{Approaches}
\label{sec:technical_section}

\subsection{Space-time baselines}

\begin{figure*}[t]
\centering
   \includegraphics[width=0.91\linewidth]{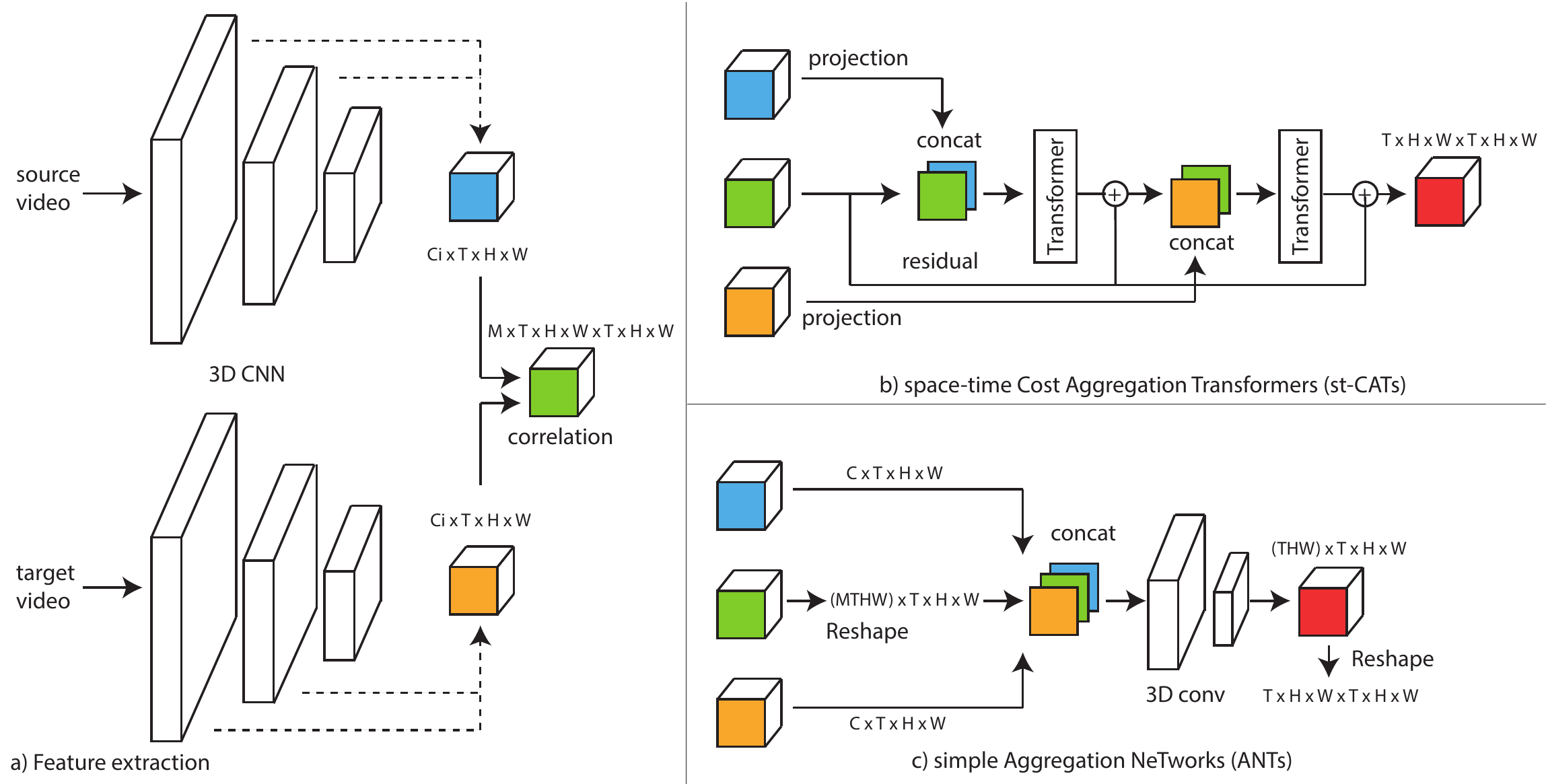}
   \vspace{-12pt}
   \caption{{\bf Baseline Approaches}. (a) both source and target videos are fed into a 3D CNN backbone for feature extraction. All selected feature maps (from different layers) are up- or down-sampled into the same feature size of T$\times$H$\times$W with their specific number of channel $C_i$. Correlation volumes are computed per feature map, then concatenated to form a tensor of M$\times$(T$\times$H$\times$W)$^2$ where M is the number of selected feature maps. (b) A space-time Cost Aggregation Transformer~\cite{cho2021cats} (st-CAT) takes both source and target feature maps and the correlation volume then applies a few transformer blocks to predict a space-time displacement flow of size (T$\times$H$\times$W)$^2$. (c) A simple Aggregation NeTwork (ANT) reshapes the correlation volume, concatenates it with the source and target features, applies a few 3D convolution layers, then predicts a space-time displacement flow of size (T$\times$H$\times$W)$^2$.}
   \label{fig:beaselines}
\end{figure*}

{\bf Overview of the approaches}. Given a pair of videos size t$\times$h$\times$w (where t is the number of frames, and h$\times$w is the frame size), the dense correspondence prediction problem can be formulated as finding a matching tensor size (t$\times$h$\times$w)$^2$~\footnote{for simplicity we denote (t$\times$h$\times$w)$^2$ instead of its full notation of t$\times$h$\times$w$\times$t$\times$h$\times$w} which encodes the matching likelihood for all pairs of pixels in the two videos. Since matching in the pixel space is costly and also less robust to semantic content, the matching is preferably done at the feature level, \eg, videos are fed into a feature extraction backbone to produce a feature map of T$\times$H$\times$W (where T, H, W are much smaller than t, h, w), to predict a smaller matching flow of (T$\times$H$\times$W)$^2$. At inference, upsampling is used to render predictions at the pixel level. Figure~\ref{fig:beaselines} presents our two baseline approaches which follow the paradigm of feature extraction followed by matching.

{\bf Feature extraction and correlation volume construction}. One common practice in the visual semantic correspondence problem in images is to extract features at different layers and then up- or down-sample the features into the same size (\eg, hyperpixel~\cite{min2019hyperpixel}). We follow the same practice but instead we use a 3D CNN backbone for videos. As shown in Figure~\ref{fig:beaselines}a, features at different layers of a 3D CNN backbone are extracted and then up- or down-sampled into the same size of T$\times$H$\times$W. The two feature maps from both source and target videos are used to construct a correlation cost volume with a size of (T$\times$H$\times$W)$^2$. We note that the correlation volume is computed independently per feature map, thus if we have M selected layers (for feature extraction), then M correlation volumes are constructed and concatenated into M$\times$(T$\times$H$\times$W)$^2$. This correlation volume and both source and target feature maps are fed into an aggregator network to produce matching predictions.

{\bf The space-time CATs}. The {\bf C}ost {\bf A}ggregation {\bf T}ransformers (CATs)~\cite{cho2021cats} is the state-of-the-art for the image visual correspondence prediction problem. Here we adopt CATs 
for our problem. We extend CATs to work on space-time feature maps of T$\times$H$\times$W instead of H$\times$W. The networks perform linear projections on the feature maps (both source and target), and sequentially concatenate the projected feature maps (source, then target) with the correlation volume. The transformers (multi-head attention) blocks and transpose are applied after each concatenation. Skip connections are also employed for stabilizing the training. The st-CATs predict a flow map size (T$\times$H$\times$W)$^2$ followed by an L2-loss w.r.t the ground truth sparse flow. Figure~\ref{fig:beaselines}b visualizes the architecture of our st-CATs baseline. For further details, readers are referred to read~\cite{cho2021cats}. We name this baseline space-time CATs to differentiate it from our later baselines where we use CATs in the space-only problem.

{\bf The simple Aggregation NeTworks}. Besides adopting CATs, we also introduce ANTs (simple {\bf A}ggregation {\bf N}e{\bf T}works) as another baseline. Similar to CATs, ANTs also take both feature maps of the source and the target videos and the correlation volume as input and predict a small flow map at the feature map size. In contrast, instead of using transformers, ANTs employ a few layers of 3D convolutions for aggregation and prediction. Our ANTs baseline is shown in Figure~\ref{fig:beaselines}c. ANTs first reshape the correlation volume, then concatenate it with both source and target feature maps. The concatenated maps are fed into a few-layer 3D CNN. ANTs use appropriate padding and no striding in its 3D convolution layers, thus the feature map size remains the same as T$\times$H$\times$W. For the final layer or prediction layer, ANTs map it back into $THW$ channels, then reshape the prediction into the size of (T$\times$H$\times$W)$^2$. The same L2 loss that was used for training CATs is applied for ANTs.

{\bf The space-time MATCH baseline}. Besides st-CATs and ANTs, we also provide a simplified version of both st-CATs and ANTs. The st-MATCH is a non-trainable version of feature matching in space-time. It takes the correlation volume (in Figure~\ref{fig:beaselines} a), and performs a mean pooling over M channels to obtain the prediction size of (T$\times$H$\times$W)$^2$.

\subsection{Sequential baselines}
One may wonder if we can decompose this problem into two sub-problems: time alignment~\cite{Dwibedi_2019_CVPR} and then spatial alignment (aka visual semantic correspondence). We present here some baselines for approaching this problem sequentially: time, then space alignment.

{\bf Time alignment options}. Visual features are extracted at each frame. We can use Nearest Neighbor (NN) search or Dynamic Time-Wraping (DTW) for alignment. Time-Cycle Consistency (TCC)~\cite{Dwibedi_2019_CVPR} is also a strong alternative for time alignment.

{\bf Space alignment options}. This step assumes that time alignment has already been done. Thus, any frame from the source video now has exactly one frame from the target video which is matched/aligned to it using one of the time alignment options. The problem is now reduced to space alignment or visual correspondence prediction: i.e., given two frames (one source and one target) and a set of keypoints in the source image, it then asks the model to predict the set of correspondences of those keypoints in the target image. Since we have many keypoints are human keypoints, one may wonder if a simple pose estimator can solve the problem. We present a pose-based baseline for space alignment. The same pose estimator~\cite{sun2019deep} is first employed to detect human poses in both source and target images. The source keypoint is used to find the closest detected human pose w.r.t the source keypoint. If there is more than one detected human pose in the target image, a simple pose descriptor is used to find the most similar pose in the target image. Finally, one of keypoints (on the matched pose in the target image) is returned as the predicted correspondence of the source keypoint. Besides the pose-based baseline, we also employ CATs~\cite{cho2021cats} as another baseline for space alignment.

\section{Experiments}
\label{sec:exp}

\subsection{Implementation details}

{\bf Setup}. We train our baseline models on the training set and evaluate them on the validation of Penn Action using 2 benchmark setups described in section~\ref{sec:benchmark}.

{\bf Input \& augmentation}. Each video (either source or target) undergoes independent augmentation. Given an input video, we randomly select a clip of 64 frames (~2 seconds at 30fps) such that it covers all key moments of that video (i.e., no keypoints are cropped). Standard image augmentations such as grayscale, posterize, equalize, rightness contrast adjusting, and solarize are all applied with a probability of $0.2$. After standard augmentations, random cropping is also applied with a probability of $0.5$. We note that all frames from the clip need to go through the same set of augmentations, otherwise the clip is no longer temporal coherent. The cropped (or uncropped) clip is then scaled to have a frame size of 128$\times$128, making the input clip a size of 64$\times$128$\times$128. Note that when random cropping and/or scaling is used, the keypoints are shifted and/or scaled accordingly.

{\bf Backbone architectures}. We experiment with different 3D CNN backbones including R3D with 18 layers and R(2+1)D with 34 layers~\cite{tran18}, both were pre-trained with Kinetics-400~\cite{kinetics}. The feature maps at different layers are either up- or down-sampled into the same size of 8$\times$8$\times$8. Due to the larger memory required for video input, it is not possible to increase this feature map size even with 32G memory GPUs. For the st-MATCH baseline, since there is no trainable parameter, the feature maps can be larger to compensate for lacking learning capacity. We find the feature map size of 32$\times$16$\times$16 is best for st-MATCH with the current 32G GPU limit.

{\bf Training details}. Training is done distributedly with 8 nodes of 8 volta GPUs each with 32G memory. A mini-batch size of 2 per GPU is used and thus making an effective batch size of 128. We follow the training schedule provided in~\cite{cho2021cats}: training is done in 100 epochs with the step learning rate schedule to be reduced (divided by 2) at epoch 70, 80, and 90. The initial learning rate is set to $1.2 \times 10^{-4}$. When full backbone finetuning is used, the initial learning rate for backbone is $1.2 \times 10^{-5}$. 

{\bf Evaluation metrics}. A predicted space-time keypoint is classified as correct when it is within a close proximity to the expected ground truth, both in space and time. Formally, a predicted keypoint $(x_{pr},y_{pr},t_{pr})$ is regarded as a correct prediction w.r.t the ground keypoint $(x_{gt},y_{gt},t_{gt})$ when $|t_{pr}-t_{gt}| \le k$ and $ \|(x_{pr},y_{pr}) - (x_{gt},y_{gt})\|_2 \le \alpha \times b$, where $\alpha$ is normally 0.1, $b$ is the of the larger size of the smallest bounding covering key-points in that frame, and $k$ is the number of frames the model is allowed to miss-align, e.g., $k$=1,3,5. The spatial metric is standard in visual correspondence prediction, regarded as PCK@0.1 (percentage of correct keypoints). Our metrics are the augmented version of PCK where we add a time-misalignment allowance, denoted as T@k-PCK@0.1.

\subsection{Baseline results}

Table~\ref{tab:main_tab} presents the space-time semantic correspondence prediction results for all baselines on two benchmark setups. All the space-time baselines use the same backbone of R(2+1)D-34. The upper table presents the results of sequential baselines while the lower reports the space-time baselines' performance. In addition to sequential baselines, we also provide an upper bound for time alignment with CATs, \eg, ground truth time alignments are given and CATs are used for spatial matching.

\begin{table}[t]
\centering
{\small
\begin{tabular}{|lc|ccc|ccc|}
        \hline
        \multicolumn{2}{|c|}{Benchmark} & \multicolumn{3}{|c|}{3+3} & \multicolumn{3}{|c|}{13+3} \\ 
        \multicolumn{2}{|c|}{Metric} & \footnotesize{T@1} & \footnotesize{T@3} & \footnotesize{T@5} & \footnotesize{T@1} & \footnotesize{T@3} & \footnotesize{T@5} \\ 
        \multicolumn{2}{|c|}{in \%} & \multicolumn{3}{|c|}{PCK@0.1} & \multicolumn{3}{|c|}{PCK@0.1} \\ 
        \hline
        \multicolumn{8}{c}{sequential baselines} \\
        \hline
        \footnotesize{NN} & \footnotesize{Pose}- & ~ & ~ & 3.2 & ~ & ~ & 8.2 \\ 
        \footnotesize{DTW} & \footnotesize{based} & ~ & ~ & 3.0 & ~ & ~ & 7.7 \\ 
        \footnotesize{TCC~\cite{Dwibedi_2019_CVPR}} & & ~ & ~ & 4.2 & ~ & ~ & 10.7 \\ 
        \hline
        \footnotesize{NN} &  & ~ & ~ & 5.9 & ~ & ~ & 13.5 \\ 
        \footnotesize{DTW} & \footnotesize{CATs} & ~ & ~ & 5.6 & ~ & ~ & 12.9 \\ 
        \footnotesize{TCC~\cite{Dwibedi_2019_CVPR}} & ~\cite{cho2021cats} & ~ & ~ & 8.1 & ~ & ~ & 17.0 \\ 
        \multicolumn{2}{|l|}{\color{gray}{groundtruth}} & \multicolumn{3}{|c|}{\color{gray}{31.0}} & \multicolumn{3}{|c|}{\color{gray}{58.9}} \\ 
        \hline
        \multicolumn{8}{c}{joint space-time baselines} \\
        \hline
        \multicolumn{2}{|l|}{st-MATCH} & 4.2 & 11.6 & 15.9 & 6.2 & 17.2 & 24.7 \\ 
        \multicolumn{2}{|l|}{{\bf st-CATs}} & \underline{19.4} & \underline{34.7} & \underline{37.7} & \underline{22.7} & \underline{48.2} & \underline{55.8} \\ 
        \multicolumn{2}{|l|}{{\bf ANTs}} & {\bf 19.9} & {\bf 35.1} & {\bf 38.1} & {\bf 24.3} & {\bf 49.9} & {\bf 57.1} \\ 
        \hline
\end{tabular}%
\vspace{-8pt}
\caption{{\bf Comparison between baselines}. Space-time correspondence prediction on two benchmark setups: 3+3 and 13+3 of pose and object-keypoints, respectively. The upper table presents sequential baselines in which the problem is approached by time aligment, then spatial correspondence prediction. The lower table presents the joint space-time baselines. Our proposed baselines, st-CATs and ANTs, significantly outperform all other baselines. st-CATs and ANTs outperform the baseline of CATs (with ground-truth time aligment provided) on the 3+3 setup while comparable with this baseline on the 13+3 setup with T@5-PCK@0.1 metric. Our experimental results suggest that it is more advantaged to approach this problem jointly in space-time rather than solving the decomposed sub-problems. Sequential baselines perform poorly on T@1 and T@3 due to challenging temporal alignment using global features, for simplicity, we omit them from the table.}
\label{tab:main_tab}}
\end{table}

{\bf Sequential baselines perform poorly}. Some observations from the sequential baselines include: (i) the pose-based baselines perform poorly, indicating that the problem should be addressed directly instead of using poses as intermediate predictions, even though many keypoints are human keypoints (\eg, in 13+3 setup); (ii) TCC~\cite{Dwibedi_2019_CVPR} is consistently better than NN and DTW as expected; and (iii) TCC~\cite{Dwibedi_2019_CVPR} with CATs~\cite{cho2021cats} performs best among sequential baselines as expected but still far below space-time baselines. 

{\bf The problem should be approached jointly in space and time}. It is interesting to see even the simple st-MATCH (with no learning capacity) outperforms all sequential baselines. This indicates that the problem should be approached jointly both in space and time rather than decomposed sub-problems. This intuitively makes sense as the decomposed problems are harder with limited context for making predictions. On one hand, for the spatial correspondence sub-problem, models have limited temporal context and no notion of motions, thus it is harder for them to predict space-time keypoints. On the other hand, for the temporal alignment sub-problem, the models often give up spatial modeling, due to long sequence inputs and the model has to focus on dense temporal predictions. Our space-time semantic correspondence prediction requires sparse predictions, normally at salient space-time keypoints. 

{\bf Simple convolutions are better than transformers on small feature maps}. When comparing learning-based methods, ANTs slightly outperform st-CATs. This can be explained as the capacity of modeling larger receptive fields of the transformers used in CATs is not crucial for small feature maps, i.e., size of 8$^3$, while a few 3D convolution layers (with $3^3$ kernels) could cover such a small receptive field. At the same time, the larger parameter size in st-CATs can cause more overfitting. Last but not least, even though both st-CATs and ANTs make predictions at low-resolution displacement flows, e.g., at $8^3$, and then perform upsampling back to $64 \times 128^2$, these models still perform reasonably well. Future work on this problem should explore the trade-offs of increasing the feature map, and prediction size for higher accuracy with more memory and computation requirements.

\subsection{Model generalization}

{\bf Different activities and keypoint types bring in different challenges}. Table~\ref{tab:detailed_tab} presents the detailed performance of our ANTs on different activities and across different types of keypoints (human vs. object). When we look at the ``all'' keypoint columns, ``golf swing'' and ``bowling'' are among the easiest while ``squats'' is the hardest activity. This can be understood by the fact that both ``golf swing'' and ``bowling'' have quite distinctive poses at key-moments while ``squats'' has the poses at the key-moments are closely similar to nearby frames (before and after the key-moment frames) making time alignment harder. We note that for the first and third key-moments of ``squats'', the motion directions and patterns are also similar to nearby frames. When we look at the ``obj'' columns, ``squats'', ``tennis forehand'', and ``baseball pitch'' are among the most challenging activities. While the ``squats'' category inherits hardness from time alignment (it has low performance across all three keypoint types), ``tennis forehand'' and ``baseball pitch'' struggle with object keypoints mainly due to the presence of small objects, e.g., the ball, and with fast motions. 

\begin{table}[t]
\centering
{\small
\begin{tabular}{|l|ccc|ccc|}
        \hline
        Metric & \multicolumn{3}{|c|}{T@1-PCK@0.1} & \multicolumn{3}{|c|}{T@5-PCK@0.1} \\ 
        Keypoint type & hum & obj & all & hum & obj & all \\
        \hline
        Baseball pitch & 20.6 & 9.5 & 18.4 & 44.1 & 14.4 & 37.9 \\
        Baseball swing & 25.2 & 17.4 & 21.7 & 54.5 & 32.3 & 44.7 \\
        Bowling & 27.5 & 44.2 & 32.0 & 48.2 & 70.9 & 54.2 \\
        Golf swing & 45.6 & 26.9 & 37.6 & 84.2 & 52.5 & 69.9 \\
        Squats & 9.9 & 6.3 & 7.2 & 25.7 & 14.8 & 17.9 \\
        Tennis forehand & 21.0 & 7.7 & 14.8 & 43.5 & 12.4 & 29.0\\
        Tennis serve & 19.0 & 16.1 & 18.0 & 39.3 & 26.1 & 34.9\\
        \hline
        All & 21.7 & 17.4 & 19.9 & 43.9 & 30.7 & 38.1\\
        \hline
\end{tabular}
\vspace{-8pt}
\caption{{\bf Detailed prediction on different actitities and keypoint types}. Our ANTs model is trained and avaluated on the 3+3 setup with the T@1 and T@5 at PCK@0.1 metrics. For activities, squats is the hardest while golf swing is the easiest. For keypoint type, object keypoints are hard in ``Baseball pitch'' and ``Tennis forehand'' due to small object, e.g., the ball, and fast motions. Object keypoints in ``Bowling'' is the easiest one due to large object size and with predictable context, e.g., the human pose at keymoments.}
\label{tab:detailed_tab}}
\end{table}

{\bf ANTs fairly generalize across keypoints}. Table~\ref{tab:cross_kps_tab} presents the performance of our ANTs trained on 3+3 or 13+3 keypoints and tested on 3+3, 13+3, and r10 setups. The r10 is a new setup denoted as the keypoints in 13+3, but not in 3+3 which is equivalent to the rest 10 types of human keypoints that are not head, left and right wrists. First, when evaluated on 13+3, the model trained on 13+3 is 7.5\% higher than the one trained on 3+3, but this is not a surprise because the model trained on 13+3 has a lot more supervision. Second, when a model is trained on 3+3, but evaluated on 13+3 and r10, performance is dropped by 3.1\% and 7.9\%, respectively. As 3+3 and r10 are two sets of non-overlapped keypoint types, an accuracy of 24.4\% on T@5-PCK@0.1, when trained on 3+3 and tested on r10, is a good one compared with other baselines (see Table~\ref{tab:main_tab}). \New{Third, when comparing the performance on different evaluation setups (3+3, r10, 13+3) with the model trained on 13+3, we observe that these results are fairly similar except that the performance on 3+3 is lower than r10, this is indicated that object keypoints are more challenging compared with human keypoints.}

\begin{table}
\centering
{\small
\begin{tabular}{|l|cc|cc|cc|}
        \hline
        Evaluate ($\rightarrow$) & \multicolumn{2}{|c|}{3+3} & \multicolumn{2}{|c|}{r10} & \multicolumn{2}{|c|}{13+3} \\ 
        Train ($\downarrow$) & T@1 & T@5 & T@1 & T@5 & T@1 & T@5 \\
        \hline
        3+3 & 19.9 & 38.1 & 12.0 & 24.4 & 16.8 & 34.2 \\
        \hline
        13+3 & \New{22.8} & \New{49.4} & \New{25.2} & \New{61.1} & 24.3 & 57.1 \\
        \hline
\end{tabular}
\vspace{-10pt}
\caption{{\bf Cross keypoint type evaluation}. ANTs models are trained on 3+3, then evaluated on 3+3, 13+3, and r10 setup. The r10 setup includes all keypoints in 13+3, but exclude those in 3+3 (r10 means the rest 10 human keypoints). }
\label{tab:cross_kps_tab}}
\end{table}

\New{{\bf ANTs and CATs generalize across datasets}. We investigate to find out if our models also work on another dataset such as Pouring. We use the models pre-trained earlier on Penn Action and further fine-tune them on Pouring. Table~\ref{tab:pouring_results} presents the results of ANTs and CATs on the Pouring dataset. Both CATs and ANTs consistently outperform the st-MATCH baseline. Due to the small size of the Pouring dataset, we repeat 3 runs of CATs and ANTs and report their mean accuracy with standard deviation. For st-MATCH, there is no learning, thus repeating experiments is not needed. We note that pre-training on Penn-Action is crucial due to the small size of the Pouring dataset. For example, CATs, without Penn-Action pretraining, \eg with an R(2+1)D-34 backbone pretrained with only K400, achieves $15.1\pm1.6$, $32.1\pm3.3$, and $42.8\pm3.0$, for T@1, T@3, and T@5, respectively. These are significant performance drops of 12--20\%.}

\begin{table}
\centering
{\scriptsize
\begin{tabular}{|c|c|c|c|c|c|}
        \hline
        Model & backbone & pretrain & T@1 & T@3 & T@5 \\
        \hline
        \scriptsize{st-} & \scriptsize{R3D-18} & K400 & 3.9 & 15.2 & 24.1 \\
        \scriptsize{MATCH} & \scriptsize{R(2+1)D-34} & K400 & 3.6 & 20.2 & 28.6 \\
        \hline
        CATs & \scriptsize{R3D-18} & PenAct & 18.6 $\pm$ 0.8 & 37.8 $\pm$ 1.0 & 55.8 $\pm$ 0.7 \\
             & \scriptsize{R(2+1)D-34} & PenAct & 27.7 $\pm$ 1.6 & 53.0 $\pm$ 1.2 & 62.2 $\pm$ 1.9\\
        \hline
        ANTs & \scriptsize{R3D-18} & PenAct & 21.3 $\pm$ 0.4 & 48.9 $\pm$ 0.7 & 63.4 $\pm$ 1.5\\
             & \scriptsize{R(2+1)D-34} & PenAct & 27.6 $\pm$ 1.3 & 57.8 $\pm$ 2.8 & 64.2 $\pm$ 1.0 \\
        \hline
\end{tabular}
\vspace{-8pt}
\caption{{\bf Results on Pouring dataset}. Both ANTs and CATs consistently outperform the st-MATCH baseline on different backbones and evaluation metrics.}
\label{tab:pouring_results}}
\end{table}

\subsection{Ablation}
\vspace{-6pt}

{\bf Different backbones}. Table~\ref{tab:backbone_ablation_tab} presents the performance of CATs and ANTs using different backbones. Both baselines have the benefit of a deeper and stronger backbone when we replace an R3D-18 with an R(2+1)D-34.

\begin{table}
\centering
{\small
\begin{tabular}{|l|c|c|c|}
        \hline
        Model & backbone & 3+3 & 13+3 \\
        \hline
        CATs & R3D-18 & 31.9 & 51.7 \\
             & R(2+1)D-34 & 37.7 & 55.8 \\
        \hline
        ANTs & R3D-18 & 31.9 & 50.9 \\
             & R(2+1)D-34 & 38.1 & 57.1 \\
        \hline
\end{tabular}
\vspace{-8pt}
\caption{{\bf CATs and ANTs with different backbones}. ANTs slightly ourperform CATs on two experimenting backbones of R3D-18 and R(2+1)D-34. All reporting results are with T@5 and PCK@0.1 metric.}
\label{tab:backbone_ablation_tab}}
\end{table}

{\bf ANTs components}. Table~\ref{tab:ant_arch_ablation_tab} presents the ablation of our ANTs components. Our observation is that increasing the number of layers in ANTs slightly improves the results at the expense of more parameters and computation. For simplicity, we set the number of layers to 2 for all other experiments with ANTs. We found that it is very important to finetune the whole backbone instead of keeping it frozen.

\begin{table}
\centering
\begin{tabular}{|l|c|c|c|}
        \hline
        Model & finetuned & \# of layers & 3+3 \\
        \hline
        ANTs & \cmark & 1 & 37.6 \\
             & \cmark & 2 & 38.1 \\
             & \cmark & 3 & 38.5 \\
             & \xmark & 2 & 21.0 \\
        \hline
\end{tabular}
\vspace{-8pt}
\caption{{\bf ANTs architecture ablation}. Results of ANTs model with R(2+1)D-34 backbone using the T@5 and PCK@0.1 metric. The number of 3D convolutional layers (not include the final prediction layer) and the option of full-finetuning backbone are ablated. While significant difference can be observed when backbone is frozen, the number of layers are less sensitive.}
\label{tab:ant_arch_ablation_tab}
\end{table}

{\bf Feature layers for hyperpixel}. In the image problem, most recent works used the hyper-pixel combination provided in~\cite{min2019hyperpixel} with a ResNet-101 backbone~\cite{He2016}. The selection is done via beam search~\cite{min2019hyperpixel}. Since our problem is for video and with a different backbone, e.g., R(2+1)D-34, we conduct an ablation to find a good combination for our hyperpixel. Here we summarize the main findings (details in appendix). As a ResNet-style architecture, R(2+1)D-34 has the following components: \texttt{conv1}, followed by 4 groups of resnet blocks. We ablate with only \texttt{conv1} and the last layer of each resnet block. Our findings are: (i) adding conv1 or feature maps from group 1 hurts performance, while adding feature maps from group $2$, $3$, $4$ helps; (ii) using two last feature maps from group $2$, $3$, $4$ provides a good trade-off of memory and computation vs. accuracy.

\section{Conclusions}
\label{sec:conclusion}

We have proposed a new task of space-time semantic correspondence prediction which requires matching and aligning semantic key-points across videos. The problem is essential in various practical applications from activity coaching and sports analysis, to robot imitation learning. We introduced two new benchmarks for this problem by adding annotations to the existing Penn Action~\cite{penn_action} and Pouring~\cite{PouringDataset} datasets. Our experiments with a set of comprehensive baselines and ablations help us gain useful insights about the problem. Some potential future directions include, but not limited to, interesting applications of space-time semantic correspondence prediction, single-shot video retrieval with explainability, and self-supervised learning with space-time cycle consistency.

\section*{Acknowledgement}
The authors would like to thank Xitong Yang for helping with the distributed traing setup and experiments.

{\small
\bibliographystyle{ieee_fullname}
\bibliography{egbib}
}

\appendix
\section{Appendices}

\subsection{Addtional qualitative results}

\begin{figure*}[t]
\centering
   \includegraphics[width=0.9\linewidth]{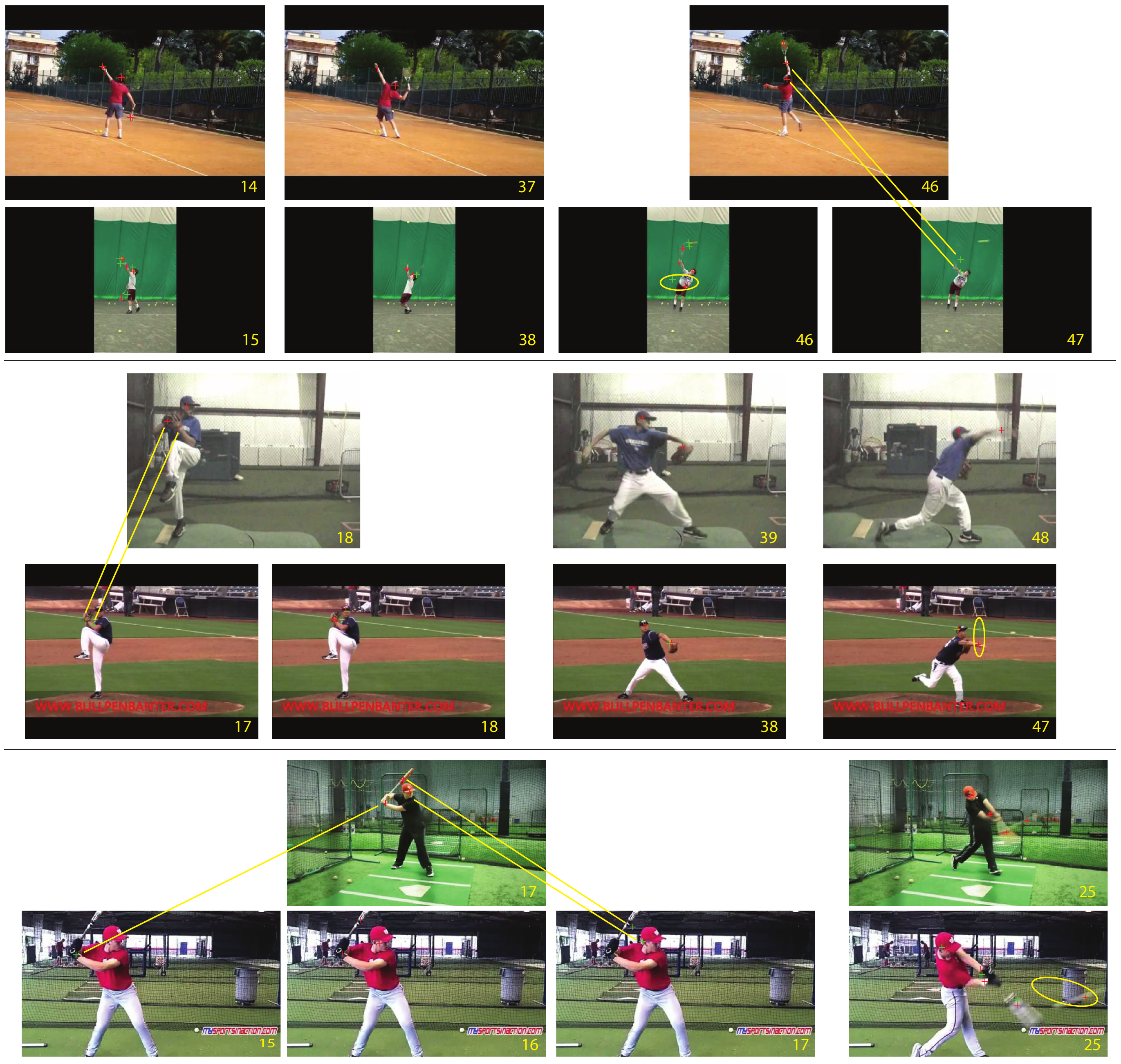}
   \vspace{-8pt}
   \caption{{\bf Qualitative results: mostly-successful cases}. Three examples of space-time semantic correspondence prediction using our ANTs: ``tennis serve'', ``baseball pitch'', and ``baseball swing''. Each example includes two rows: the upper row shows the source video, while the lower one shows the target video. Frame numbers are shown at the lower-right corner of each frame. Ground truth keypoints are visualized in red, predicted keypoints are green. Yellow lines across source and target videos indicate missed time-alignments. Yellow ellipses indicate missed space-alignments. Best viewed in color.}
   \label{fig:qual_results}
\end{figure*}

\begin{figure*}
\centering
   \includegraphics[width=\linewidth]{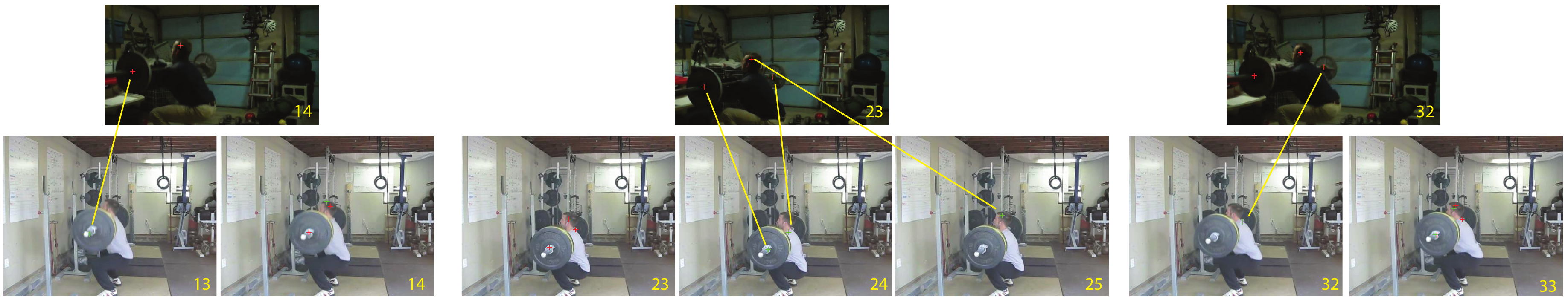}
   \vspace{-8pt}
   \caption{{\bf Qualitative results: a failure case}. The notations are similar as what used in Figure~\ref{fig:qual_results}. Our most challenging class is ``squats'' where the model failed to align in time due to similar spatial keypoint and motion pattern before and after keymoment frames. Best viewed in color.}
   \label{fig:qual_failure}
\end{figure*}

Addtional visualizations of our ANTs model predictions (on Penn Action 3+3 setup) are shown in Figure~\ref{fig:qual_results} and Figure~\ref{fig:qual_failure}. We observe that most of spatial miss-alignments happen due to small objects and / or fast motions (the ball in baseball and tennis sequences). The most challenging cases for time alignment come from ``squats'' as a result of indistinguishable motion patterns with the near-by frames, \eg, before or after the keymoment frames.

\subsection{Hyperpixel ablation}

\begin{table}
\centering
{\small
\begin{tabular}{|l|l|c|}
        \hline
        {\bf Setup} & {\bf Hyperpixel} & {\bf T@5} \\
        \hline
        Base & $\{0,3,7,13,16\}$ & 37.33 \\
        \hline
        $-\{0\}$     & $\{3,7,13,16\}$ & 38.13 \\
        $-\{3\}$     & $\{0,7,13,16\}$ &  38.26 \\
        $-\{7\}$     & $\{0,3,13,16\}$ & 38.23 \\
        $-\{13\}$     & $\{0,3,7,16\}$ & 38.17 \\
        $-\{16\}$     & $\{0,3,7,13\}$ & 37.16 \\
        \hline
        $-\{0,3\}$,$+\{12,15\}$     & $\{7,12,13,15,16\}$ & {\bf 38.38} \\
        $+\{12,15\}$     & $\{0,3,7,12,13,15,16\}$ & 38.28\\
        \hline
\end{tabular}%
\caption{{\bf Hyperpixel ablation}. Results of ANTs with an R(2+1)D-34 backbone on the 3+3 setup using different sets of hyperpixels. We find out that removing early layers, \eg, 0, 3, 7, improves accuracy while removing deeper layers, \eg, 16, degrades performance. Using two layers from from the last 2 ResNet groups is enough for a good performance.}
\label{tab:ablate_hyperpixel}
}

\end{table}
Table~\ref{tab:ablate_hyperpixel} presents the results of our ANTs with an R(2+1)D-34 backbone trained and evaluated on the 3+3 setup using different sets of hyperpixels. As a ResNet-style architecture, R(2+1)D-34 has the follwing components: \texttt{conv1}, followed by 4 groups of resnet blocks. We ablate with only \texttt{conv1} and the last layer of each resnet blocks. Thus, we have the following feature layers from our R(2+1)D-34 backbone, for simplicity we number them from 0: 0 (\texttt{conv1}), 1 to 3 (from group 1), 4 to 7 (from group 2), 8 to 13 (from group 3), and 14 to 16 (from group 4). We start with the base combination which includes the \texttt{conv1} feature and the last layer of each group (presented in the first row of Table~\ref{tab:ablate_hyperpixel}). We then ablate by removing each feature map from the base combination for sensitive analysis. We observe that removing feature maps from early layers helps improving performance, while removing deeper layers, \eg, 16, degrades accuracy. Finally, we add more layers from group 3 and 4 (the last row) or further remove early layers (the second-last row). We find out that using two last feature maps from group 3 and 4 provides a good trade-off of memory and computation vs. accuracy.

\end{document}